\documentclass[11pt,a4paper]{article}
\usepackage[hyperref]{acl2020}
\usepackage{times}
\usepackage{latexsym}
\usepackage{booktabs}

\usepackage{graphicx}

\usepackage{microtype}

\aclfinalcopy 

\title{Chart-to-Text: Generating Natural Language Descriptions for Charts by Adapting the Transformer Model}

\author{Jason Obeid \\
School of Information Technology \\
  York  University, Canada\\
  \texttt{jobeid98@my.yorku.ca} \\\And
  Enamul Hoque \\
School of Information Technology \\  
  York  University, Canada\\
  \texttt{enamulh@yorku.ca} \\}

\date{}

\begin{document}
\maketitle
\begin{abstract}
Information visualizations such as bar charts and line charts are very popular for exploring data and communicating insights. Interpreting and making sense of such visualizations can be challenging for some people, such as those who are visually impaired or have low visualization literacy. In this work, we introduce a new dataset and present a neural model for automatically generating natural language summaries for charts. The generated summaries provide an interpretation of the chart  and convey the key insights found within that chart. Our neural model is developed by extending the state-of-the-art model for the data-to-text generation task, which utilizes a transformer-based encoder-decoder architecture. We found that our approach outperforms the base model on a content selection metric by a wide margin (55.42\% vs. 8.49\%) and generates more informative, concise, and coherent summaries.

\end{abstract}

\section{Introduction}

Information visualizations such as bar charts and line charts are commonly utilized by people to get insights from data and make informed decisions. However, understanding and getting insights from charts can be difficult and time-consuming. This is because people often need to visually compare between several graphical marks (e.g. bars) of a chart to infer key insights from data, which may be challenging when the chart involves many data items~\cite{kim-visqa-2020}.

Generating a natural language summary to explain a chart has numerous benefits and potential applications. It can help users in understanding and interpreting charts by conveying key points about the chart by focusing on temporal, causal, and evaluative aspects~\cite{carenini2013user}. It can help people identify
insights from charts that they otherwise might have missed. A previous study conducted on a chart corpus found that the text associated with the chart failed to convey any insight from that chart in 35\% of the instances, while in another 26\% of cases the text conveyed only a portion of the chart's intended message~\cite{carberry2006information}. Therefore, effective chart summarization could help data analysts, business analysts, or journalists in better preparing reports from data. 
Such summaries could also enable people who are visually impaired or have low cognitive abilities to comprehend charts and perform analytical tasks through audio~\cite{Ferres-accessibility-2013}.
 
However, natural language generation (NLG) systems for charts are still in their infancy. 
Early work mostly focused on statistical methods for finding salient information from charts and planning-based approaches for structuring content~\cite{reiter2007architecture} to generate textual captions from basic charts~\cite{fasciano1996postgraphe, mittal-etal-1998-describing, green2004autobrief,  demir-etal-2012-summarizing}. Commercial systems such as Quill~\cite{quill} and Wordsmith~\cite{wordsmith} have similarly adopted statistical algorithms and template-based NLG methods which are used to produce data facts in textual form. 
Unfortunately, the predefined template-based NLG methods and planning-based architecture for generating summaries often lack generality and may not offer variations in style. Moving beyond template and planning-based approaches~\cite{reiter2007architecture},
more recently researchers considered data-driven neural models for generating text from data tables~\cite{mei2016talk, li2019enhanced}. However, they do are not designed for chart summarization as they do not consider the chart specific features (e.g. chart type). To our knowledge, there have not been any  efforts to develop deep neural models that  are specifically tailored for generating chart summaries.

\begin{figure}[t!]
 \includegraphics[width=\linewidth]{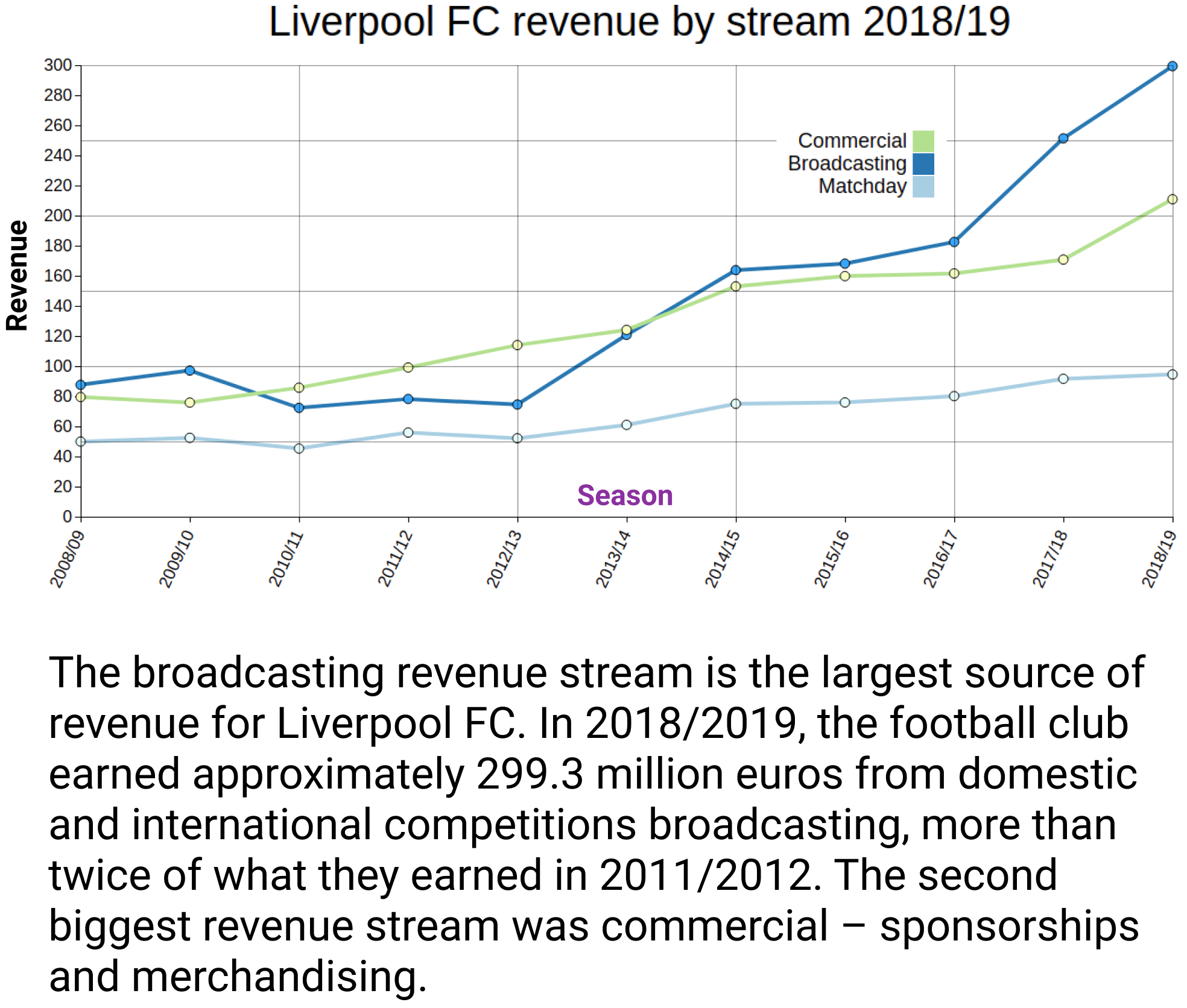}
\caption{
An example of a generated natural language explanation from a chart from our model.
}
\label{fig:example-chart}
 \end{figure}
 
In this paper, we present a neural model for 
chart summarization by 
extending a transformer-based model that was originally designed for the data-to-text generation task. A key challenge in developing neural models for chart summarization is the lack of a suitable dataset containing a large set of charts and corresponding human-written summaries. To address the challenge we first developed a corpus of 8,305 charts, where for each chart we crawled the corresponding data table and the human-written summary. 
Our model learns content selection and text generation from such a collection of chart-summary pairs to generate 
summaries of charts. We also introduce a data variable substitution method that replaces the tokens referring to chart data with  variables to 
generate more factually correct statements than a competitive baseline. As shown in Figure \ref{fig:example-chart}, the resulting summary is quite informative, concise, and coherent.

The contributions of our paper are three-fold:
     \vspace{-2mm}
\begin{itemize}
     \item First, we introduce a new large-scale corpus on chart summarization consisting of human-written summaries of charts along with chart images and their underlying data.
     \item Second, we adopt a transformer-based  model to generate chart summaries, which learns from chart-summary pairs from our dataset. To our knowledge, our work is the first to investigate the problem of chart summarization using a data-driven deep neural model.
     \item Finally, we perform a series of evaluations to compare our model's performance with a baseline model derived from Gong et al. \cite{li2019enhanced}. As a secondary contribution, we will make our source codes and the new dataset used in this research publicly available.
\end{itemize}

\section{Related Work}
\subsection{Chart Summarization}
Early work on generating natural language summarization follows some planning-based approaches which can be captured by Reiter's NLG architecture for data-to-text generation~\cite{reiter2007architecture}.
For example,~\cite{mittal-etal-1998-describing} developed a caption generation system which determines the structure of the caption based on the mappings between the data and marks of the charts (such as text, lines, and bars) and chooses a complexity metric to select details of the caption. It then uses a text planner to generate the description of the chart. The iGRAPH-Lite system~\cite{Ferres-accessibility-2013} aims to make charts accessible to blind users via generating captions and supporting navigation through the chart using a keyboard. Like ~\cite{mittal-etal-1998-describing} it uses templates to provide a short description of what the chart looks like.
Nonetheless, these systems describe the chart only in terms of how to interpret the chart rather than explaining high-level insights conveyed by the chart. 

Other research has focused on generating multimedia presentations combining both text summary and charts~\cite{Fasciano2000IntentionsIT, green2004autobrief}. PostGraphe takes user-specified intention (e.g. increasing trend) and a spreadsheet file as input and then generates a simple chart and caption separately using some heuristics~\cite{Fasciano2000IntentionsIT}. Autobrief generates presentations in text and information graphics~\cite{green2004autobrief} using an integrated planning-based approach in the domain of transportation scheduling. 

 
 
 


There has also been growing interest to automatically extract insights from a dataset and present them using template-based NLG. Examples of such automatic insight generation include research prototypes such as DataSite~\cite{cui2019datasite}, DataShot~\cite{wang2019datashot} and Voder~\cite{srinivasan2018augmenting} as well as commercial systems like Quill\footnote{Narrative Science. \href{https://narrativescience.com/quill/}{Quill}}, Arria\footnote{Arria, \href{https://www.arria.com/business-intelligence/}{Arria}}, and Wordsmith\footnote{Wordsmith \href{https://automatedinsights.com/wordsmith}{Wordsmith}}. These systems typically perform some statistical analysis to infer potentially important or interesting facts about the data and then present them in natural language sentences and charts. ~\cite{demir-etal-2012-summarizing} present a method that computes some statistics (e.g. min, max, trends) and identifies the intended message that a bar chart is conveying using a Bayesian inference system. Then, it generates the summary in a bottom–up approach to simultaneously construct the discourse and sentence structures of textual summaries. More recently, \cite{DBLP:journals/corr/abs-1906-02850} uses the encoder-decoder architecture where a Residual Network (ResNet) \cite{he2016deep} in the encoder is used to recognize the input chart from an image, and Long Short-Term Memory (LSTM) and attention in the decoder to create template-based captions. 

A common limitation of the above body of work is that the sentences are generated
using predefined template-based approaches which may
lack generality and
offer fewer variations in grammatical style and lexical choices compared to data-driven models.
In contrast, we focus on learning language variations and automatically finding important insights from charts using a deep learning model on a large collection of chart-summary pairs.

\subsection{Data-to-Text Generation}
The objective of data-to-text generation is to generate a descriptive summary given structured data. Data-to-text generation focuses on creating a descriptive summary from structured data which can be encoded as a table of records. Data to text generation has been explored for various domain-specific tasks such as summarizing sport game data~\cite{barzilay-lapata-2005-collective,  liang2009learning, wiseman2017challenges}, weather-forecast data
~\cite{reiter2005choosing}, recipe generation~\cite{yang2017reference} and biography generation~\cite{lebret-etal-2016-neural}. 

Several recent methods have primarily focused on using sequence-to-sequence learning methods \cite{mei2016talk, lebret-etal-2016-neural, wiseman2017challenges}. For example,~\cite{mei2016talk} propose an encoder-decoder model that uses recurrent neural networks with LSTM units for jointly learning content selection and surface realization for weather forecast and soccer commentaries.  \cite{wiseman2017challenges} present a new dataset on  NBA game summarization and evaluate several models including attention-based encoder-decoder models. They found that neural text generation techniques from data perform poorly at content selection and lack inter-sentential coherence. \cite{puduppully2019data} attempt to address this problem by incorporating content selection and planning mechanisms within the neural model. \cite{li2019enhanced} found that the transformer model yielded outputs more fluent and coherent when compared to their seq2seq counterparts, which is why we use it as the base model. 

\section{Chart Summarization Dataset 
}
There have been several benchmark datasets that are made available recently for the data-to-text generation tasks \cite{lebret-etal-2016-neural, wiseman2017challenges,  chen2020logical, parikh2020totto}. However, to the best of our knowledge, there are no publicly available large datasets of chart data paired with human-generated summaries. While some prior work on generating captions and summaries from charts (e.g.  ~\cite{mittal-etal-1998-describing, green2004autobrief, demir-etal-2012-summarizing}) exist, to our knowledge they do not provide any large datasets with chart-summary pairs.

With the above challenges in mind, we create a new dataset for chart summarization which is available at \url{https://github.com/JasonObeid/Chart2Text}. In order to find the best source of data for our corpus, we have analyzed publicly available charts from various sources such as textbooks, research papers, news articles, and websites that contain data charts and facts. This led us to choose Statista\footnote{https://www.statista.com/} as our data source. Statista regularly publishes charts from data collected by market and opinion research institutes, and data derived from the economic sector. Since Statista has all the necessary metadata including the data tables, titles, and concise summaries of charts it was a suitable source for our purpose. The dataset was crawled from 23,382 freely accessible pages from statista.com in early March of 2020, yielding a total of 8,305 charts, and associated summaries. For each chart, we downloaded the chart image, the underlying data table, the title, the axis labels, and a human-written summary describing the statistic.

After examining the crawled dataset, we found that out of 8,305 charts, 7,726 charts were lacking the x-axis labels. To address this problem we searched by regular expressions and applied named entity recognition using CoreNLP~\cite{manning-EtAl:2014} on the data table to automatically identify if the x-axis represented common temporal dimensions such as years or months. For the remaining 1,353 missing labels we used human annotators who examined the title, summary and the chart to come up with short descriptive labels for the x-axis.

\begin{table}[t!]
\centering
\begin{tabular}{@{}lccc@{}}
\toprule
               & Line            & Bar       & Total:   \\ \midrule
    Simple     & 3564            & 3199      & 6763     \\
    Complex    & 902             & 640       & 1542     \\ \bottomrule
\multicolumn{1}{r}{Total:} & 4466            & 3839     \\
\end{tabular}
\caption{Chart type distribution}
\vspace{-3mm}
\label{tab:dataset-statistics2}
\end{table}

\begin{table}[t!]
\centering
\begin{tabular}{@{}lc@{}}
\toprule
    Statistic           & Value  \\ \midrule
    Mean Token Count    & 113.4   \\
    Mean Sentence Count & 5.2    \\
    Vocab Size          & 19,150 \\
    Total Tokens        & 941.8k\\
    Mean Data Cells     & 32.3 \\ \bottomrule
\end{tabular}
\caption{Dataset statistics}
\vspace{-3mm}
\label{tab:dataset-statistics1}
\end{table}

The resulting dataset can be described as follows: let the set of chart-summaries $CS = [I, D, T, L, S]$, where for each chart-summary in $CS$ there is a chart image $i \in I$, data table $d \in D$, title $t \in T$, axis labels $l \in L$, and summary $s \in S$. The dataset consists of both bar and line charts (Table \ref{tab:dataset-statistics1}). A simple bar chart only has a set of bars, and a simple line chart contains a single line. Complex bar charts include stacked and grouped bar charts while complex line charts have multiple lines. 

The summaries of the charts are concise, with an average sentence count of 5.2, and an average token count of 113.4. The data tables were moderate in size with an average of 32.3 cells. The discourse structures of these summaries were usually similar. They 
usually start by describing the chart at a high-level in terms of what this chart is about and then describing and comparing salient points in the chart. Some common salient information and statistics  are extremes (e.g. highest/lowest values) or trends (upward/downward tendencies) or simple value retrieval (e.g. mentioning the first/last data point's value).



\section{The Chart-to-Text Model}

\begin{figure*}[t!]
 \includegraphics[width=\linewidth]{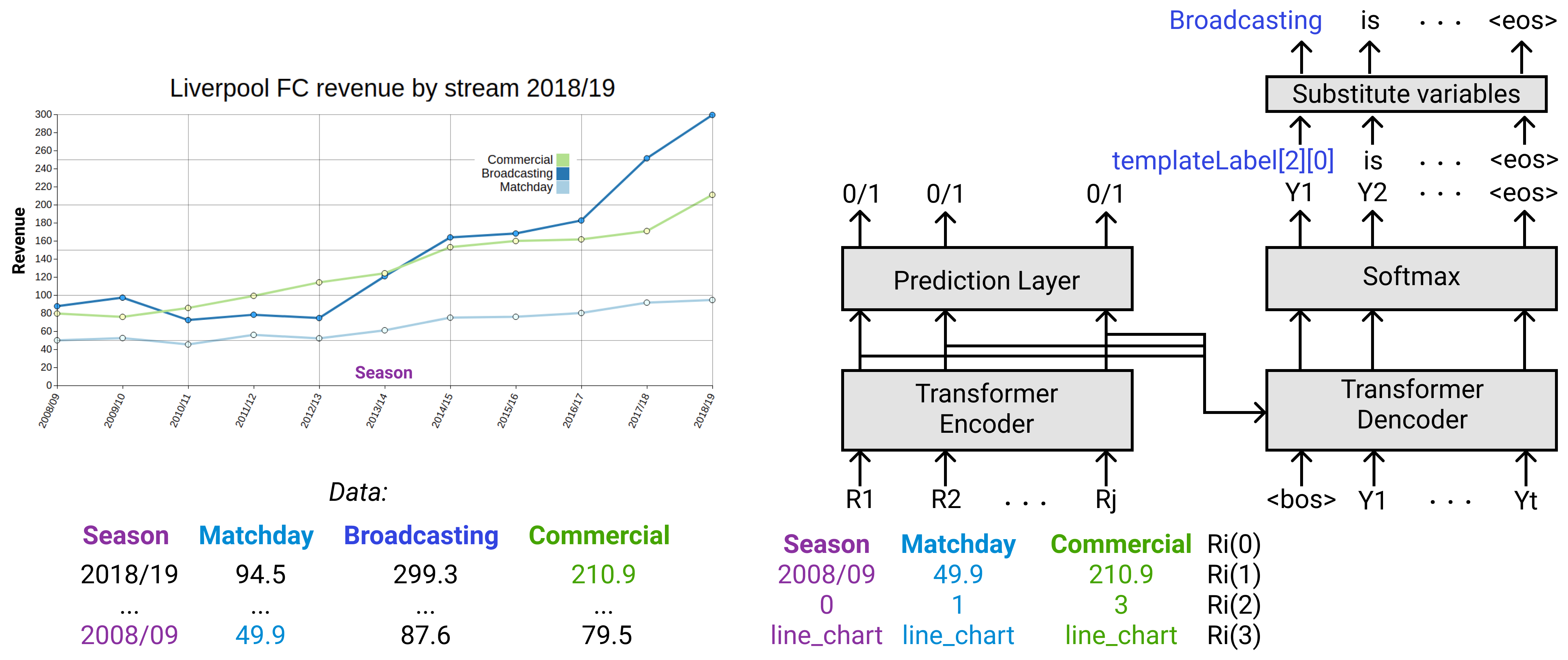}
\caption{
An overview of the proposed approach for chart summarization using a transformer-based model. The model takes the data table and some chart metadata as input (on the left) and generates a summary containing data variables that refer values within the data table.
}
\label{fig:ModelOverview}
 \end{figure*}
 

In this section, we first describe the Transformer-based Model for Data-to-Text \cite{li2019enhanced} which we use as our base model, followed by our adaptations to this model for generating chart summaries. Then, we describe our experiments with  model parameters and the training procedure.


\subsection{Base Model}
Our base model \cite{li2019enhanced} extends the standard transformer \cite{vaswani2017attention} by adding a binary prediction layer and a content selection training step. The input layer of the model accepts a tuple of four features (entity, type, value, information) as input. The model then generates the latent representation of each record in the input sequence and passes it through the binary prediction layer. The decoder of this model is the same as the original Transformer model and predicts the summary based on encoder's output.
The model also removes the positional embedding of the transformer encoder, as there is no ordered relationship within the records of their dataset. 



\subsection{Our Proposed Approach}

Figure~\ref{fig:ModelOverview} shows an overview of our proposed approach. The model takes data records and other chart information as input. Like the  
base model~\cite{li2019enhanced},  it uses the self-attention mechanism of the transformer encoder to generate the latent representations of the input, and the binary prediction at the top of the Transformer encoder output to decide whether or not a record will be mentioned in the target summary. Finally, the decoder predicts the next token in the context of the encoder output and the previous tokens from the summary sequence.

In order to adapt the enhanced transformer model~\cite{li2019enhanced} for generating chart summarization, we introduce three main changes. First, we modify the four features of the record tuples used as input to the model. This is done to include additional information which is important for chart summarization. Second, we reintroduce positional embeddings to the encoder since charts tend to contain ordered relationships. 
Finally, we introduce a process of substituting tokens with data variables to minimize the amount of generated hallucinations, a problem commonly found in NLG where a predicted word is not grounded in truth \cite{wiseman2017challenges, parikh2020totto}. We now discuss each of these changes in more details:



\textbf{Input embedding:}
For each data table $d \in$ D, we pre-process it into a set of records $r \in R$. Each of these records are placed in a tuple with four features as follows:
\begin{itemize}
\vspace{-2mm}
     \item r\textsubscript{i}(0) contains the column header,
     \vspace{-3mm}
    \item r\textsubscript{i}(1) contains the table cell value,
    \vspace{-3mm}
    \item r\textsubscript{i}(2) contains the column index, and 
    \vspace{-3mm}
    \item r\textsubscript{i}(3) contains the chart type
\end{itemize}
Each feature is embedded into a vector, and together they are concatenated to represent the record as shown below:

$\forall$ r $\in$ R:

$r\textsubscript{i} = [r\textsubscript{i}(0); r\textsubscript{i}(1); r\textsubscript{i}(2); r\textsubscript{i}(3)$]

Our model takes each of these records r\textsubscript{i} as input, and outputs a set of predicted tokens which we will denote as y $\in$ Y. To get the final summary, for each token y\textsubscript{i} in y, if the token is a data variable then it is substituted with the corresponding data value.





\textbf{Positional embedding:} Unlike the sports dataset used by~\cite{li2019enhanced}, chart data often involves an ordered dimension such as temporal data (e.g. year, month), or ordinal categories in bar charts. In order to generate summaries that better capture salient features from such sequential data, we re-introduce positional embeddings to our model. 

\textbf{Data variable substitution:} A critical issue with various existing models for data-to-text generation problem such as ~\cite{li2019enhanced} is that they treat data records mentioned in the summaries as regular tokens which often results in hallucination problem. As a consequence, the model sometimes predicts tokens that are irrelevant to the chart and thus results in factually incorrect sentences. This problem becomes even more serious for our chart dataset which is not focused on a specific domain (e.g. sports). In order to improve the factual accuracy of the generated summaries we introduce a data variable substitution method. 

\begin{figure}[t!]
 \includegraphics[width=\linewidth]{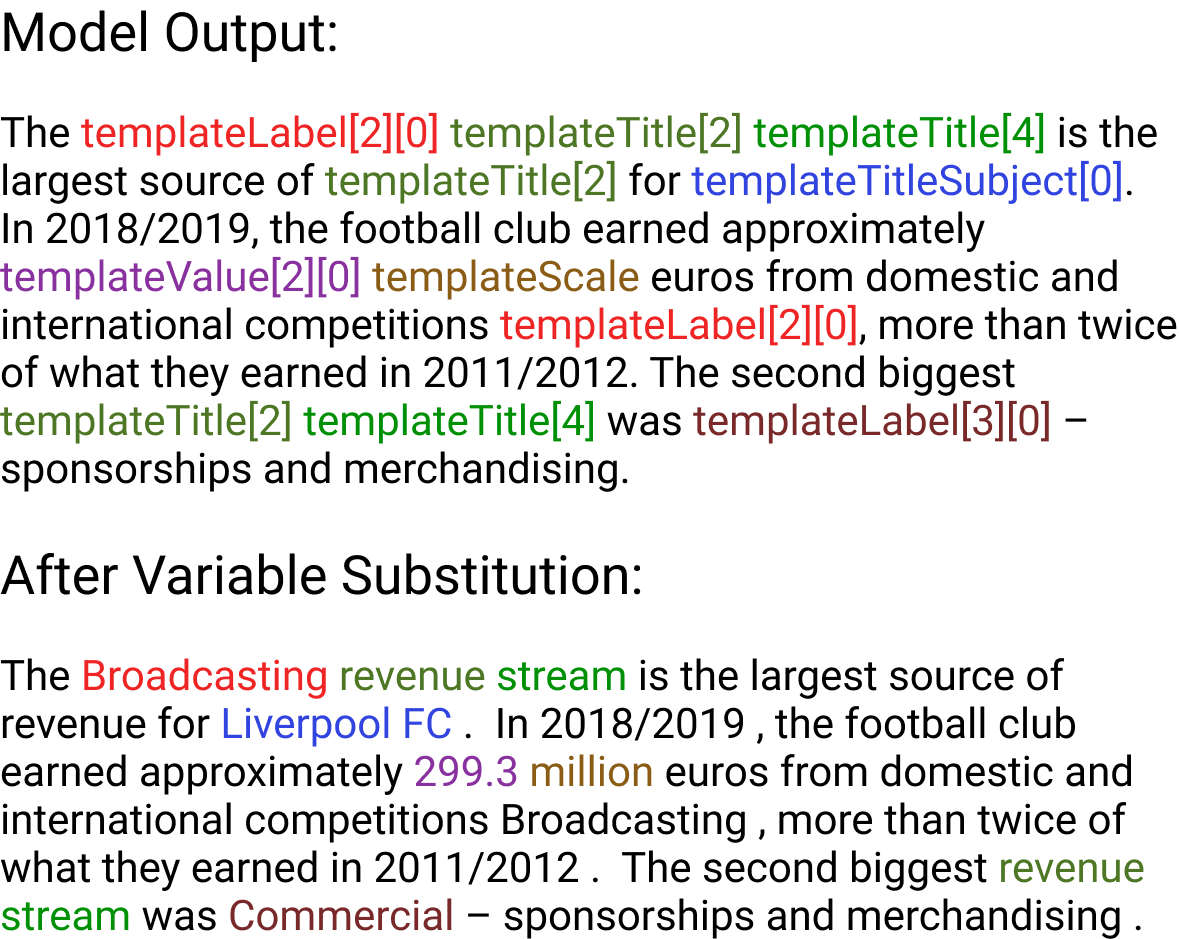}
\caption{
Demonstration of data variable substitution.
}
\vspace{-2mm}
\label{fig:variableSubstitution}
 \end{figure}

The idea is that before training we first modify the gold summaries so that whenever a token references something from the data table, chart title, or axis labels we replace them with one of the predefined data variables. We then use these modified summaries to train the model so that it learns how to generate the summary more generally with data variables as opposed to actual values in the data table or tokens from the title. Then, during the testing phase, if a generated token matches a predefined data variable, we make a look-up operation to convert the data variable into the referenced chart data (see Figure ~\ref{fig:variableSubstitution}). 

We define seven categories of data variables listed in order of priority: subjects, dates, axis labels, titles, table cells, trends, and scales. Here subjects refer to entities relevant to the data table (e.g. `Liverpool FC') and trends refer to tokens that indicate positive or negative trends (e.g. growing, decreasing, etc). Scales refer to tokens such as `millions' and `percentage' that are associated with numeric tokens. For each token $t\textsubscript{i}$ in a gold summary $s$, if that token matches any of these variable categories, then it is substituted by that variable. We apply named entity recognition~\cite{manning-EtAl:2014}  to detect the subjects and dates. The rest of the variables are detected using simple string pattern matching techniques. For each variable, we also assign the relevant index position in the data table or other chart data. For example, given a sentence in the gold summary ``Broadcasting is the largest source of revenue for Liverpool FC'', the system replaces the token `Broadcasting' with \emph{$<templateLabel[2][0]>$} which refers to the third column's header label. 

During the testing phase, our approach performs the reverse operation where it substitutes data variables with their referenced tokens. Figure~\ref{fig:variableSubstitution} demonstrates how the data variables are substituted with actual tokens from the chart data. Here, the corresponding tokens and variables are highlighted by the same color.



\subsection{Training}
The model requires two types of labels for its training step. The first type of label is for the chart's data, which for each record r\textsubscript{i} in the set of records $r$ there is either a label of $1$ if $r\textsubscript{i}$ is mentioned in the summary $s$, or a 0 if not. The second type of label is for the chart's summary, where for each token t\textsubscript{i} $\in$ a gold summary $s$ there is a label of $1$ if $t$ is also present in some records in $r$, or a $0$ otherwise.

The dataset was divided into training, validation, and testing sets with a 70\%:15\%:15\% ratio for training. We trained the model for 80 epochs with an epoch size of 1000, using the following hyper-parameters: 1 encoder layer, 6 decoder layers, embedding size of 512, batch size of 6, beam size of 4, sinusoidal positional embeddings, and GELU \cite{DBLP:journals/corr/HendrycksG16} activations.
\section{Evaluation}
We first perform an automatic evaluation to approximately measure the performance of our approach compared to the base model~\cite{li2019enhanced} and then we conduct a human evaluation. Finally, we perform some qualitative analysis to have a better understanding of the effectiveness and trade-offs of our approach.


\subsection{Automatic Evaluation} 

For automated evaluation of our summary quality, we employ two different metrics: (1) the BLEU score \cite{papineni2002bleu}, and (2) a content selection metric inspired by \cite{wiseman2017challenges}. We calculate our content selection metric as the percentage of records mentioned in the generated summary that are mentioned in the gold summary. 

Table~\ref{tab:automated-eval1} shows the results of automatic evaluation of our model compared to the base model. We observe that our model slightly improves over the baseline on the BLEU metric and outperforms it on content selection by a wide margin. In particular, the huge improvement on the content selection metric suggests that by learning how to generate summaries in terms of data variables rather than actual data values, our approach largely addresses the hallucination problem.

\begin{table}[]
\centering
\begin{tabular}{@{}lcc@{}}
\toprule 
\multicolumn{3}{r}{Summarization Method}\\\midrule
Evaluation Method & Our Model  & Gong et al. \\\midrule 
BLEU Score        & 18.54             & 17.06 \\
Content Selection & 55.42 (24.06) & 8.49 (9.99) \\\bottomrule
\end{tabular}
\caption{ Results of automatic evaluation for two summarization methods (The content section measure is shown in percentage and the standard deviation is mentioned in the bracket).}
\label{tab:automated-eval1}
\end{table}

\subsection{Human Evaluation}
To further investigate the quality of the generated text, we perform
a human evaluation. For this purpose, we randomly sample 40 different charts where we have 10 charts from each of four chart types (simple bar, complex bar, simple line, complex line). We use the model from \cite{li2019enhanced} trained on our dataset with positional embeddings enabled as our baseline. We created a Mechanical Turk study and surveyed three unique respondents per statistic. We use the survey to assess the quality of each summary from four independent perspectives: (1) \textbf{Informativeness}: How informative is the summary of the chart? (2) \textbf{Conciseness}: How concise is the summary of the chart?, (3) \textbf{Coherence}: How coherent is the summary of the chart? and (4) \textbf{Fluency}: How fluent or grammatically correct are the sentences in the summary of the chart? The questions were asked using a 5-point Likert scale from 1 (the worst) to 5 (the best).

As shown in Table~\ref{tab:human-eval1}, on average our model outperforms the base model in terms of informativeness, conciseness, and coherence by at least over 1 point. Overall, it indicates 
that our method can generate summaries that are more insightful, that have better connections between sentences and with fewer repetitions. In terms of fluency, the base model performs slightly better (3.78) than our model (3.73).


\begin{table}[]
\centering
\begin{tabular}{@{}lcc@{}}
\toprule 
\multicolumn{3}{r}{Summary Method}\\\midrule
Evaluation Category & Our Model  & Gong et al. \\\midrule
Informativeness & 3.42  (1.16) & 2.13  (1.44)\\
Conciseness     & 3.58  (1.11) & 2.40  (1.47)\\
Coherence       & 3.32  (1.25) & 2.27  (1.43)\\
Fluency         & 3.73  (1.07) & 3.78  (0.85)\\\bottomrule
\end{tabular}
\caption{Results of human evaluation (standard deviation in brackets)}
\label{tab:human-eval1}
\vspace{-2mm}

\end{table}


We also wanted to evaluate the factuality aspect of the generated summaries. For this purpose, we asked respondents to evaluate whether the facts stated in each sentence of the summary were supported by the chart. There were four possible responses to this question: yes, no, partially, and can't decide (explain why). Table \ref{tab:human-eval2} shows the percentage of statements within the summary that were perceived as factually correct or not. We find that over 50\% sentences generated by our model were verified as factually correct, whereas only 22\% sentences from the baseline model's summaries were perceived to be correct. Another 15.56\% statements generated from our model were verified as partially correct compared to 7.24\% from the base model. This result suggests that our model generates more factually correct statements compared to the base model.



\begin{table}[]
\centering
\begin{tabular}{@{}lcc@{}}
\\\toprule 
\multicolumn{3}{r}{Summary Method}\\\midrule
Response & Our Model  & Gong et al. \\\midrule 
Yes           & 51.02\% & 22.70\% \\
No            & 26.30\% & 66.02\% \\
Partial       & 15.56\% & 7.24\% \\
Can't decide         & 7.13\%  & 4.04\% \\\bottomrule
\end{tabular}
\caption{Responses for factual correctness}
\label{tab:human-eval2}
\vspace{-2mm}
\end{table}


\begin{figure}[t!]
 \includegraphics[width=\linewidth]{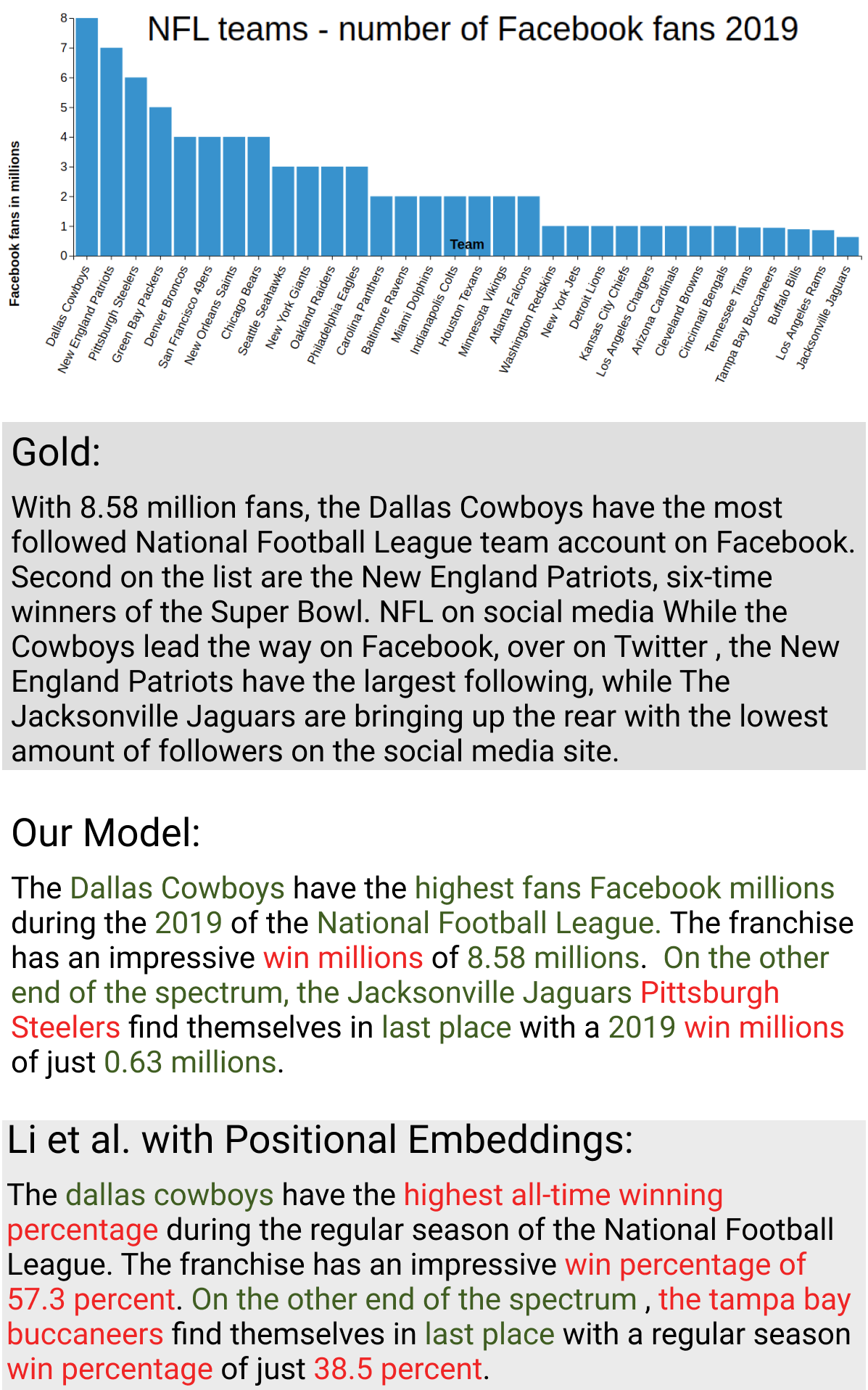}
\caption{
Comparison of summary generation methods. Here, factually correct statements are highlighted in green and incorrect statements are highlighted in red.
}
\label{fig:case_study}
 \end{figure}
\subsection{Case Study}
In order to better understand how our approach works we reviewed one random sample from our dataset (see Figure \ref{fig:case_study}). We notice that the gold summary is relatively longer than the two generated summaries with some additional external facts that are not mentioned in the chart data (e.g. `the New England Patriots' and `their Twitter following'). This illustrates the challenge for generating chart summaries because without direct connections between the chart and its summary, it becomes more difficult for the model to learn what is relevant. Our model's summary is similar in structure to the gold, but without mention of these external statistics. The summary is mostly accurate, as it correctly mentions the most and least followed teams along with their corresponding fan counts (highlighted in green), but it incorrectly predicts tokens related to game wins, and also incorrectly mentions the third highest team (highlighted in red) right after it correctly mentioned the lowest team. When analyzing the summary generated from  \cite{li2019enhanced}, we can readily see that it makes more false predictions than our model. It correctly mentions the first team name, but then after that 
it suffers from the hallucination problem as it repeats a memorized summary which it was trained on. As a result, this summary is 
mostly irrelevant to the given chart. This is also reflected from the automatic evaluation (Table \ref{tab:automated-eval1}) where the model from Gong et al. performed poorly on the content selection metric.

\begin{figure}[t!]
 \includegraphics[width=\linewidth]{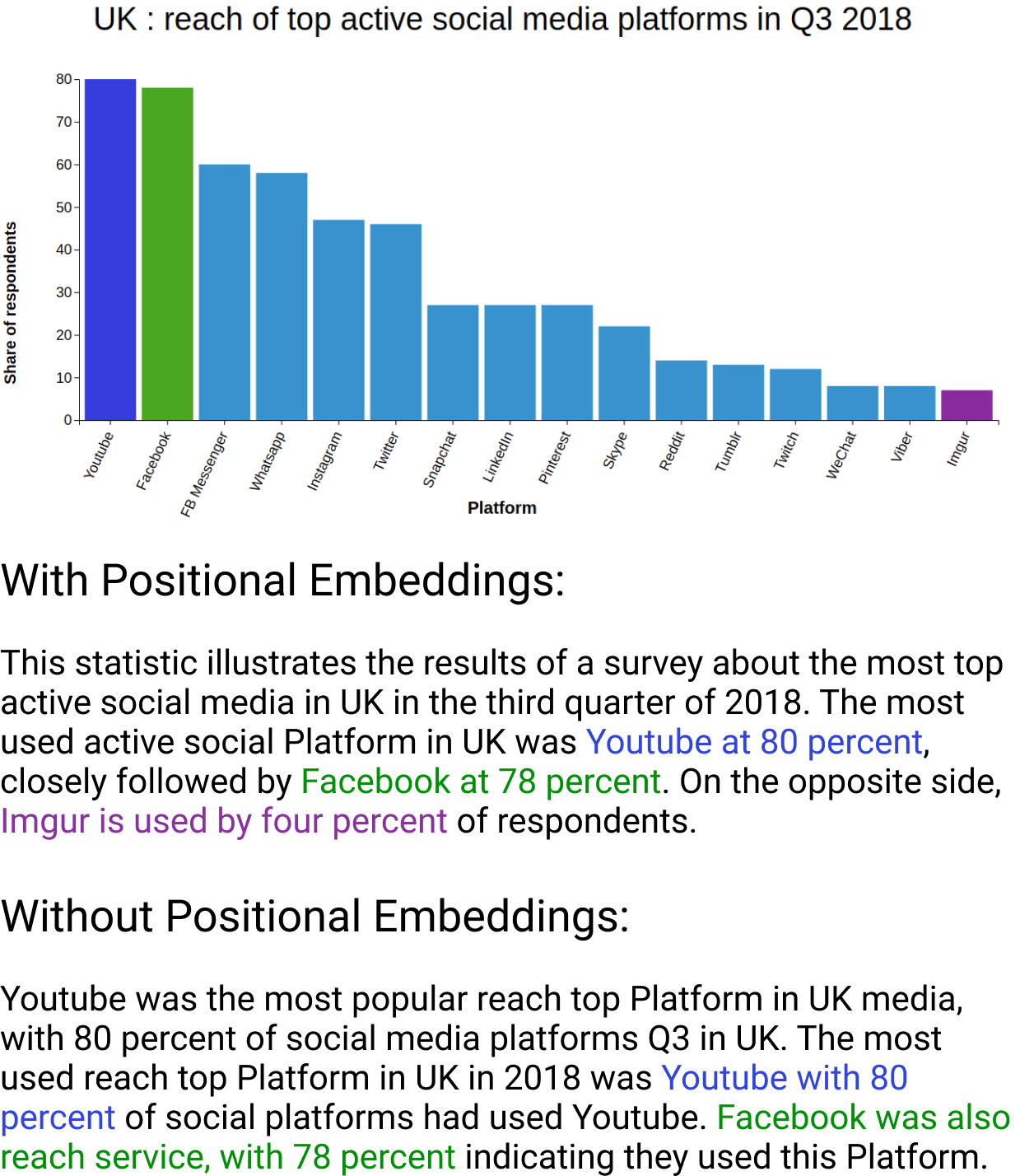}
\caption{
Comparison of generated summaries with positional embeddings enabled and disabled (Here the text that refers to a bar in the chart is highlighted with the same fill-color of that bar).
}
\vspace{-4mm}
\label{fig:position-embedding}
 \end{figure}

We also investigated the impact of positional embeddings on our model's performance. As we can see from Figure \ref{fig:position-embedding}, for a sample chart the model that enables positional embeddings yielded summaries with a more meaningful discourse structure by better connecting between sentences and with more meaningful salient information. This support that positional embeddings allow the model to track sequential data as  suggested by \cite{vaswani2017attention} for the original Transformer model.


\begin{figure}[t!]
 \includegraphics[width=\linewidth]{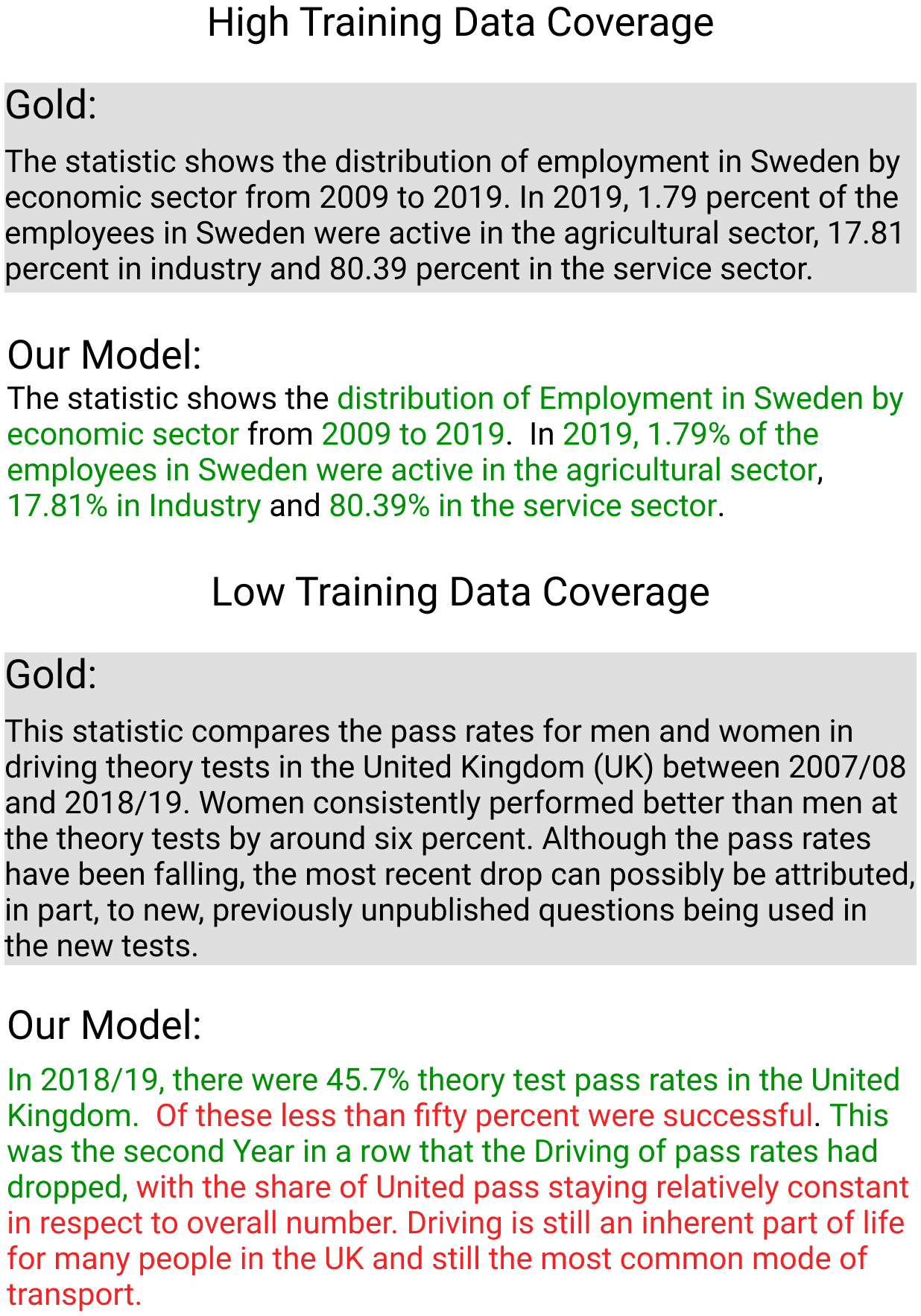}
\caption{
Comparison of generated summaries by our model based on data coverage. 
}
\vspace{-2mm}
\label{fig:dataCoverage}
 \end{figure}
\subsection{Error Analysis}
While our evaluation reveals that our model is more effective at chart summary generation than the base model, it still has much room for improvement. In particular, we identify two categories of common errors which are present in our generated summaries. The most common error is the fact hallucination, which is demonstrated in Figure \ref{fig:case_study}. While our model has largely addressed this issue, irrelevant tokens still occasionally occur. Our error analysis on 50 random samples indicates that hallucination can occur in two main ways in our model. First, sometimes the model predicts a data variable at an incorrect index, for example it may try to refer to one index of a table row, but the correct index is actually that of another index.  The second hallucination method is when the model simply predicts a token which is irrelevant to the data, which we see commonly in other data-to-text works  \cite{wiseman2017challenges, parikh2020totto}. This error usually occurs when the model generates summaries of charts about a domain that has low coverage in the training set. For example, our training set has several charts on finance and sports. For these charts, the generated summaries are quite fluent and tokens  that are irrelevant to the corresponding chart are rare. Figure~\ref{fig:dataCoverage} shows how the generated summaries differ in fluency and amount of irrelevant tokens for two different charts: one from a domain with low training data coverage and the other with high training data coverage.

\section{Conclusion and Future Work}
In this paper, we tackle the challenge of automatic chart summarization by introducing  a new dataset  and proposing a neural approach based on the transformer architecture. Our approach learns how to generate natural language descriptions from charts by modifying the gold summaries to replace references to chart data values with data variables. As a result, the model learns 
how to summarize in a more generalized way,
and generates more factually correct statements compared to the base model. Our model also generates summaries that are more informative, concise, and coherent according to human evaluations. We hope that our work will motivate researchers to further improve the quality of summaries automatically generated from charts, which is a highly under-explored research area.

In the future, we would like to develop larger datasets that cover more diverse domains and additional chart types (e.g. pie charts, scatterplots, heatmaps etc.) to further improve the quality and generalizability of our model. Also, while we compared our method with a strong baseline from  \cite{li2019enhanced}, we will perform further comparisons with other models which were developed for the data-to-text generation problem. Finally, 
we would like to build applications such as an 
interactive chart summarization system with a focus on enhancing the accessibility of charts so that blind and visually impaired people can comprehend charts via audio.

\section*{Acknowledgments}

This work was supported by the Natural Sciences and Engineering Research Council (NSERC), Canada. We thank
Dhruv Nayyar for helping with preparing the dataset.

\bibliography{acl2020}
\bibliographystyle{acl_natbib}

\end{document}